\documentclass{article}

\usepackage[nonatbib,preprint]{neurips_2022}

\usepackage[utf8]{inputenc} 
\usepackage[T1]{fontenc}    
\usepackage{hyperref}       
\usepackage{url}            
\usepackage{booktabs}       
\usepackage{amsfonts}       
\usepackage{nicefrac}       
\usepackage{microtype}      
\usepackage{xcolor}         

\usepackage{multirow}
\usepackage{colortbl}
\usepackage{subfigure}

\usepackage{algpseudocode}
\usepackage[ruled,vlined]{algorithm2e}

\usepackage{amsmath}
\usepackage{amssymb}
\usepackage{mathtools}
\usepackage{amsthm}

\usepackage{tcolorbox}
\tcbuselibrary{most}
\newtcolorbox{mybox}[2][]{colbacktitle=red!10!white, colback=gray!10!white,coltitle=black!70!black, title={#2},fonttitle=\bfseries,#1}

\definecolor{commentcolor}{RGB}{110,154,155}   
\newcommand{\PyComment}[1]{\ttfamily\textcolor{commentcolor}{\# #1}}  
\newcommand{\QuadPyComment}[1]{\ttfamily\textcolor{commentcolor}{\quad \# #1}}  
\newcommand{\PyCode}[1]{\ttfamily\textcolor{black}{#1}} 

\title{The Ladder in Chaos: A Simple and Effective Improvement to General DRL Algorithms by \\ Policy Path Trimming and Boosting}

\author{%
  Hongyao Tang, Min Zhang, Jianye Hao \\
  College of Intelligence and Computing\\
  Tianjin University\\
  \texttt{\{bluecontra,min\_zhang,jianye.hao\}@tju.edu.cn} \\
}

\begin{document}

\maketitle

\begin{abstract}
Knowing the learning dynamics of policy is significant to unveiling the mysteries of Reinforcement Learning (RL).
It is especially crucial yet challenging to Deep RL, from which the remedies to notorious issues like sample inefficiency and learning instability could be obtained.
In this paper, we study how the policy networks of typical DRL agents evolve during the learning process by empirically investigating several kinds of temporal change for each policy parameter.
On typical MuJoCo and DeepMind Control Suite (DMC) benchmarks, we find common phenomena for TD3 and RAD agents:
1) the activity of policy network parameters is highly asymmetric and policy networks advance monotonically along very few major parameter directions;
2) severe detours occur in parameter update and harmonic-like changes are observed for all minor parameter directions.
By performing a novel temporal SVD along policy learning path,
the major and minor parameter directions are identified as the columns of right unitary matrix associated with dominant and insignificant singular values respectively.
Driven by the discoveries above, we propose a simple and effective method, called \textbf{Policy Path Trimming and Boosting (PPTB)}, as a general plug-in improvement to DRL algorithms.
The key idea of PPTB is to periodically trim the policy learning path by canceling the policy updates in minor parameter directions,
while boost the learning path by encouraging the advance in major directions.
In experiments, we demonstrate the general and significant performance improvements brought by PPTB, when combined with TD3 and RAD in MuJoCo and DMC environments respectively.
\end{abstract}

\section{Introduction}
\label{sec:intro}

Deep Reinforcement Learning (DRL) has achieved a lot of impressive results in different sequential decision-making problems, e.g., video game~\cite{BadiaPKSVGB20Agent57}, robot navigation~\cite{Shah22LMNav}, mathematics~\cite{FawziBHHRB0RSSS22AlphaTensor}, nuclear fusion control~\cite{DegraveFBNTCEHA22NuclearControl} and chatbot~\cite{chatgpt}.
Despite the great potential of DRL demonstrated by these achievements,
DRL is far from well understood as convergence and learning dynamics of DRL agents remain mysterious at present.
This impedes the development of more advanced algorithms and also prevents the deployment of DRL agents in broader real-world scenarios.

Although the learning dynamics of RL agents has been studied with tabular~\cite{Sutton1988ReinforcementLA} and linear approximation~\cite{Ghosh2020RepresentationsFS}, knowing the learning dynamics of DRL agents is challenging.
The difficulty comes from the complex interplay between Deep Learning models and DRL algorithms, and further escalates when only limited online interactions or offline logged data are considered.
Recently, there are a few works that study the learning dynamics of DRL agents from different perspectives.
A major stream of works among them focus on \textit{Co-Learning Dynamics} between representation and RL functions (usually the value network)~\cite{DabneyBRDQBS21VIP,KumarAGL21ImplicitUnder,LyleRD22Understanding,NikishinSDBC22PrimacyBias,Tang2022SelfPred}.
The focus of this stream is that DRL agent over-shapes its representation towards early experiences and objectives (e.g., TD targets of early policies)
while becomes less capable for later learning process.
This degradation becomes more severe and detrimental gradually due to the non-stationary learning nature, finally leading to myopic convergence or even divergence.
From another angle, 
a phenomenon called \textit{Policy Churn} is discovered in~\cite{Schaul22PolicyChurn}. It reveals that the greedy policy induced by a typical value network changes its actions on about 10\% states after only a single update.
A very recent~\cite{Sokar2302Dormant} work presents a phenomenon called \textit{Dormant Neuron} in DRL: the neurons of typical value networks gradually become inactive during the learning process, leading to the loss of expressivity.
Different remedies are proposed to address the corresponding issues concerned in these works.

In this paper,
we aim at unveiling the learning dynamics of policy network over the training course of typical DRL agents.
We resort to a key angle called \textit{policy learning path}, i.e., the evolvement history of a policy network.
Taking MuJoCo~\cite{Brockman2016Gym} and DeepMind
Control Suite (DMC)~\cite{Tassa2018DMC} as typical DRL benchmarks,
we conduct a series of empirical investigations on the policy learning paths of TD3~\cite{Fujimoto2018TD3} and RAD~\cite{LaskinLSPAS20RAD} agents respectively.
To analyze policy learning path,
we make use of several measures of temporal change for each policy network parameter.
Moreover, we propose a novel approach called \textit{Temporal SVD} to view the evolvement of policy network parameters in a low-dimensional space spanned by the columns of right unitary matrix associated with significant singular values.
We summarize four commonly observed phenomena from our empirical investigations.
To be concrete,
the accumulated absolute change amount of policy network parameters is highly unbalanced and many parameters have very small changes throughout the learning process;
meanwhile, severe detours exist in parameter update.
From the lens of temporal SVD,
we observe that singular value information is highly concentrated on the first a few singular values.
By viewing the evolvement of policy networks along the rows of left unitary matrix,
we find that 
policy networks advance monotonically along only very few major parameter directions and show harmonic-like oscillations on all other minor parameter directions.
In addition, we show that policy networks can be reconstructed at negligible performance loss only with singular value information of the first few major parameter directions.

This drives us to ask a question:
\textit{can we make the DRL agent focus on the policy learning along major directions while suppress the oscillation on minor directions?}
To this end, we propose a simple method called \textbf{Policy Path Trimming and Boosting (PPTB)}, as a general plug-in improvement to DRL agents.
The key idea of PPTB is to trim the policy learning path by canceling the policy parameter updates
in minor dimensions occasionally, while boost the learning path by encouraging the advance in major directions.
To incorporate PPTB into a typical DRL algorithm, only a few codes need to be added to realize the maintenance of (proximal) policy learning path, parameter modification with temporal SVD and reloading.
Finally, we evaluate the effects of PPTB based on TD3 and RAD in several MuJoCo and DMC environments.

Key contributions of this work are summarized below:
\begin{itemize}
    \item We conduct a series of empirical study on policy learning path of typical DRL agents.
    
    \item We summarize four common phenomena which reveals the asymmetry and detour of policy parameter updates, and the distinct evolvement behaviors on major and minor parameter directions.

    \item We propose a simple and easy-to-implement improvement for general DRL algorithms. The effectiveness of our proposed method is demonstrated in popular MuJoCo and DMC environments based on TD3 and RAD.
\end{itemize}

\section{Preliminaries}
\label{sec:prelim}

\paragraph{Markov Decision Process (MDP)}
Consider a Markov Decision Process (MDP) $\left< \mathcal{S}, \mathcal{A}, \mathcal{P}, \mathcal{R},  \gamma, \rho_0, T \right>$,
defined with a state set $\mathcal{S}$, an action set $\mathcal{A}$, the transition function $\mathcal{P}: \mathcal{S} \times \mathcal{A} \times \mathcal{S} \rightarrow \mathbb{R}$, the reward function $\mathcal{R}: \mathcal{S} \times \mathcal{A} \rightarrow \mathbb{R}$, the discounted factor $\gamma \in [0,1)$, the initial state distribution $\rho_0$ and the horizon $T$.
The agent interacts with the MDP by performing its policy $\pi: \mathcal{S} \rightarrow P(\mathcal{A})$ that defines the distribution over all actions for each state.
The objective of an RL agent is to optimize its policy to maximize the expected discounted cumulative reward
$J(\pi) = \mathbb{E}_{\pi} [\sum_{t=0}^{T}\gamma^{t} r_t ]$,
where $s_{0} \sim \rho_{0}\left(s_{0}\right)$, $a_{t} \sim \pi\left(s_{t}\right)$, $s_{t+1} \sim \mathcal{P}\left(s_{t+1} \mid s_{t}, a_{t}\right)$ and $r_t = \mathcal{R}\left(s_{t},a_{t}\right)$.
The state-action value function $Q^{\pi}$ is defined as 
$Q^{\pi}(s, a)=\mathbb{E}_{\pi} \left[\sum_{t=0}^{T} \gamma^{t} r_{t} \mid s_{0}=s, a_{0}=a \right]$
for all $s,a \in \mathcal{S} \times \mathcal{A}$.

\paragraph{Deep Reinforcement Learning (DRL)}

With function approximation based on deep neural networks, an RL agent is able to deal with large and continuous state-action space.
Conventionally, $Q^{\pi}$ can be approximated by $Q_{\phi}$ with parameters $\phi$
typically through minimizing Temporal Difference loss \cite{Sutton1988ReinforcementLA}, 
i.e., 
$L(\phi)=\frac{1}{2}\left[Q_{\phi}(s, a) - \mathbb{E}_{a^{\prime} \sim \pi(s^{\prime})} \left(r + \gamma Q_{\bar{\phi}}(s^{\prime},a^{\prime})\right) \right]^{2}$.
A parameterized policy $\pi_{\theta}$, with parameters $\theta$, can be updated by taking the gradient of the objective, 
i.e., $\theta^{\prime} \leftarrow \theta + \eta \nabla_{\theta} J(\pi_{\theta})$ with a learning rate $\eta$.
Therefore,
starting from a initial policy $\pi_{\theta_1}$,
the DRL agent proceeds along learning process and obtains a sequence of policies $\{ \pi_{\theta_1}, \pi_{\theta_2}, \dots \}$.
For two representative DRL algorithms, 
Deterministic Policy Gradient (DPG) theorem \cite{Silver2014DPG} is often used to update a deterministic policy with the gradient:
$\nabla_{\theta} J(\pi_{\theta}) = \mathbb{E}_{s \sim \rho^{\pi_{\theta}}} \left[ \nabla_{\theta} \pi_{\theta}(s) \nabla_{a} Q_{\phi}(s,a)|_{a=\pi_{\theta}(s)}\right]$,
where $\rho^{\pi_{\theta}}$ is the discounted state distribution under policy $\pi_{\theta}$;
Soft Actor-Critic (SAC)~\cite{HaarnojaZAL18SAC} updates a stochastic policy with the gradient:
$\nabla_{\theta} \hat{J}(\pi_{\theta}) = \mathbb{E}_{s \sim \rho^{\pi_{\theta}}} \big[ \nabla_{\theta} \log \pi_{\theta}(a|s) + (\nabla_{a} \log \pi_{\theta}(a|s)) - \nabla_{a} Q_{\phi}(s,a)) \nabla_{\theta} f_{\theta}(\epsilon ;s)) |_{a=f_{\theta}(\epsilon ; s)}\big]$ (with noise $\epsilon$ and implicit function $f_{\theta}$ for re-parameterization),
based on the maximum-entropy objective $\tilde{J}(\pi_{\theta}) = \mathbb{E}_{\pi} [\sum_{t=0}^{T}\gamma^{t} r_t + \xi \mathcal{H}(\pi_{\theta}(\cdot|s)]$ where $\xi$ is the temperature of entropy term.




\section{Phenomena on DRL Policy Learning Path}
\label{sec:phen}

In this section, we conduct empirical investigations on typical DRL policy learning process. In specific, we use official codes of TD3~\cite{Fujimoto2018TD3} and RAD~\cite{LaskinLSPAS20RAD} for OpenAI MuJoCo~\cite{Brockman2016Gym} continuous control environments and visual-input DeepMind Control (DMC) Suite~\cite{Tassa2018DMC} environments respectively.
Both of them use a policy architecture of two-layer MLP (where the output layer is viewed as \textit{Layer 3}). Note that we exclude the convolution layers for image representation in RAD here and focus on the policy part~\cite{ChungNJW19TwoTimescale}.
All the experimental details can be found in Appendix~\ref{appsec:additional_results}.

First of all, we call the sequence of policies $\{\pi_1, \pi_2, \dots, \pi_n \}$ or policy parameters $\{\theta_1, \theta_2, \dots, \theta_n \}$ obtained during the learning process of a typical DRL algorithm as \textit{policy learning path} or \textit{policy parameter path} throughout this paper. Similar concepts are also used in~\cite{DabneyBRDQBS21VIP,tang2022PeVFA}. In the following, we focus on how policy network parameters evolve along policy learning path from a few angles.
Based on the insightful phenomena discovered in this section, we propose a simple plug-in method to improve general DRL algorithms in Sec.~\ref{sec:ppt_method}.


\subsection{Policy Parameter Change and Detour}
\label{subsec:param_change}

\begin{figure*}
\begin{center}
\subfigure[MuJoCo/Hopper]{
\includegraphics[width=1.0\textwidth]{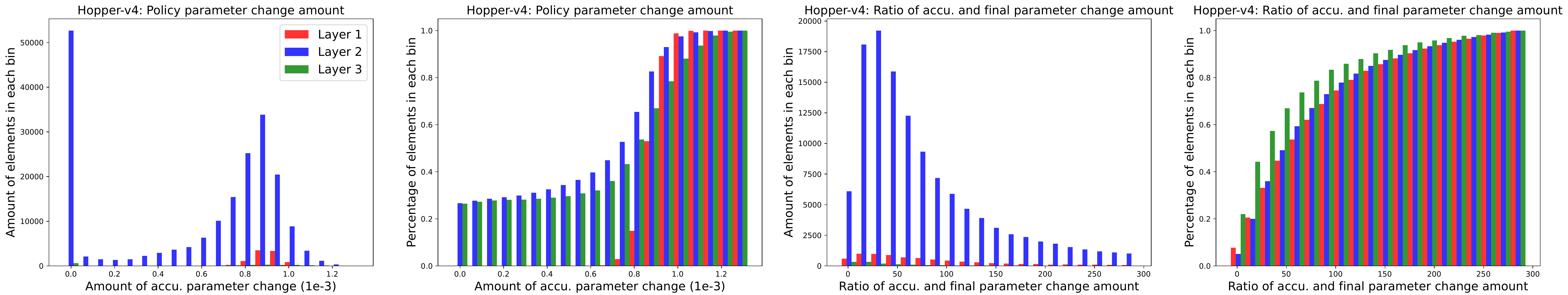}
}\\
\subfigure[DMC/walker-walk]{
\includegraphics[width=1.0\textwidth]{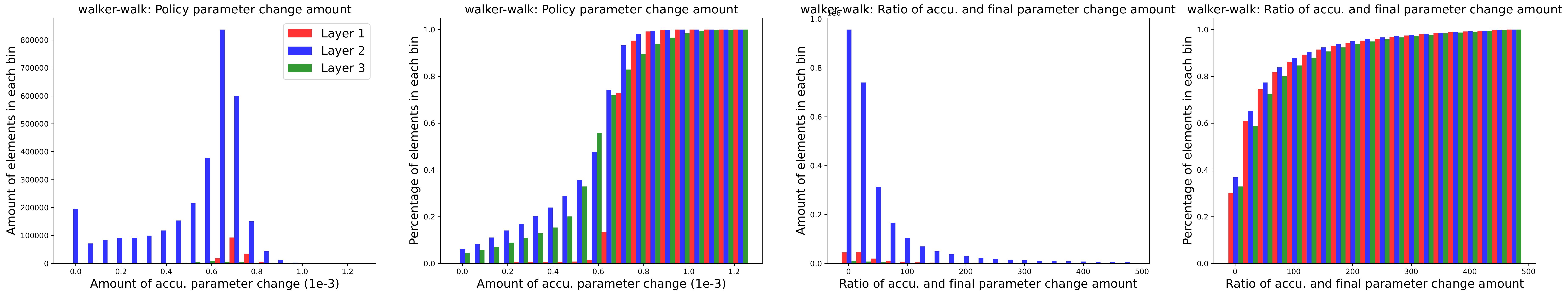}
}
\end{center}
\caption{\textbf{Policy Parameter Change and Detour} on policy paths obtained by TD3 on MuJoCo/Hopper and RAD on DMC/walker-walk. 
Each row contains: (\textit{left 1}) number histogram of accumulated absolute change amount ($\Delta_{\text{apc}}$) of each parameter and (\textit{left 2}) its corresponding cdf histogram;
(\textit{right 2}) number histogram of parameter update detour ratio ($r_{\text{rup}}$) of each parameter and (\textit{right 1}) its corresponding cdf histogram.
Only upper 80\% parameters according to $\Delta_{\text{apc}}$ are taken to plot their $r_{\text{rup}}$ for meaningful analysis;
extreme values of $r_{\text{rup}}$ are neglected for clarity.
See \textbf{Phenomenon 1.1} and \textbf{2.1} for conclusions.
}
\label{figure:ana_param_change}
\end{figure*}

We start from tracking the absolute change amount of each parameter in policy network.
First, we introduce a few quantities below for our following investigations:
\begin{itemize}
    \item Accumulated Parameter Change: $\Delta^{\text{apc}}(\{\theta_i\}_{i=1}^{n}) = \sum_{i=1}^{n-1}|\theta_{i+1} - \theta_{i}|$.
    \item Final Parameter Change:
    $\Delta^{\text{fpc}}(\{\theta_i\}_{i=1}^{n}) = |\theta_{n} - \theta_{1}|$.
    \item Parameter Update Detour Ratio:
    $r^{\text{pud}}(\{\theta_i\}_{i=1}^{n}) = \frac{\Delta^{\text{apc}}(\{\theta_i\}_{i=1}^n)}{\Delta^{\text{fpc}}(\{\theta_i\}_{i=1}^n)}$.
\end{itemize}
We use $\Delta^{\text{apc}},\Delta^{\text{fpc}},r^{\text{pud}}$ as abbreviations when the context is clear.
For each environment, we run TD3 or RAD for three trials and collect the policies along the learning process at intervals.
We then calculate $\Delta^{\text{apc}},r^{\text{pud}}$ for each policy parameter and plot the histograms in a layer-wise manner (i.e., Layer 1,2,3).
Note that the parameter number of Layer 2 is often much higher than those of Layer 1,3.
The results are shown in Fig.~\ref{figure:ana_param_change}.
The results are similar in almost all other MuJoCo and DMC environments and the complete figures can be found in Appendix~\ref{appsec:additional_results_change_amount}.

We first check the amount of policy parameter change. According to the histograms of $\Delta^{\text{apc}}$ (see the first two plots of each row), we observe our first phenomenon.

\begin{mybox}[detach title,before upper={\tcbtitle\quad}]{Phenomenon 1.1 (\textit{Parameter Change Asymmetry}):}
There is a significant discrepancy in the change amount among policy parameters. The second and the output layer have similar patterns which differ from that of the first layer.
\end{mybox}

A certain proportion (about 10\% - 40\%, varying among different environments) of parameters in the second and the output layer have minor changes, while the parameter change of first layer is relatively even.
One thing to note is that the results reported in Fig.~\ref{figure:ana_param_change} are based on the policy paths of entire learning process, thus being a global view. 
For a more local view (i.e., a window of recent learning process), the phenomenon of parameter update asymmetry becomes more obvious and more parameters have minor and even no change (i.e., dead parameters).
We hypothesize that this could be some practical evidence of lazy training of neural network~\cite{Chizat2018OnLazyTraining}.
Concretely, the phenomenon indicates that a lot of parameters have insignificant gradients $\nabla_a Q(s,a)\nabla_{\theta}\pi_{\theta}(s) |_{a=\pi(s)}$ during learning process.

Next, we concern to what extent each policy parameter detours from its initial value to its final value. According to the histograms of $r^{\text{pud}}$ (see the last two plots of each row), we observe our second phenomenon.

\begin{mybox}[detach title,before upper={\tcbtitle\quad}]{Phenomenon 2.1 (\textit{Parameter Update Detour}):}
There are severe detours in policy parameter update. All the three layers show similar patterns.
\end{mybox}

It can be easy to consider that such detours or oscillations in policy parameter updates can be mainly attributed to noisy policy gradients. This also reveals the correlation between policy parameters where most of them can not be updated independently.

Although few previous works have made a comprehensive quantitative investigation on how policy network parameters evolve in typical DRL learning process,
the two phenomena introduced above are not very surprising.
More importantly, we do not gain useful insights that help in improving DRL learning process till now.
Thus, we go on and conduct further investigations in the following.

\begin{figure*}
\begin{center}
\subfigure[Singular Value Information Curves]{
\includegraphics[width=0.75\textwidth]{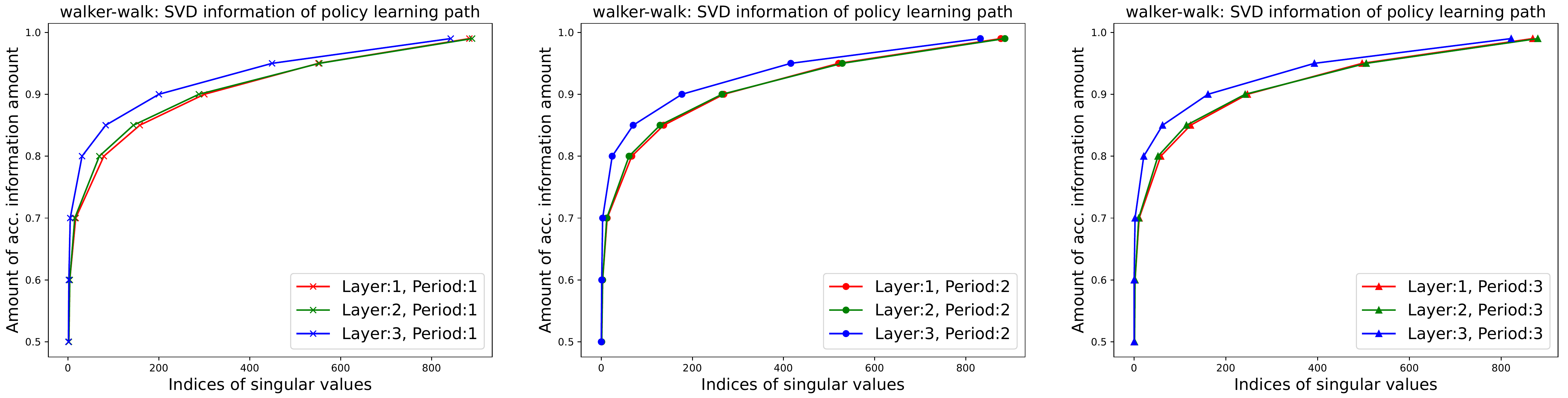}
\label{subfig:svd_curves_dmc_walker_walk}
}\\
\subfigure[SVD Left Unitary Matrix ($U$) Investigations]{
\includegraphics[width=1.0\textwidth]{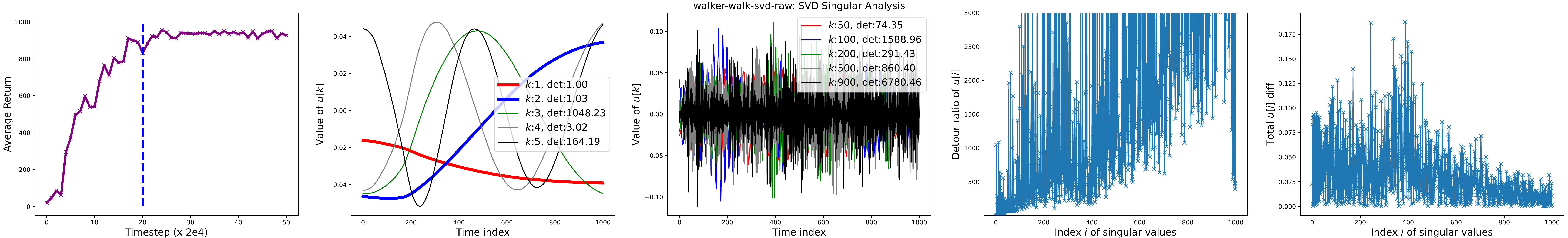}
\label{subfig:svd_u_ana_dmc_walker_walk}
}
\end{center}
\caption{\textbf{Temporal SVD Analysis} of policy paths obtained by RAD on DMC/walker-walk.
\textit{(a)} We plot the curves of $\mathbb{D}(\beta)$ against threshold $\beta \in \{0.5, 0.6, 0.7, 0.8, 0.85, 0.9, 0.95, 0.99\}$ for the three layers and three periods (i.e., early, middle, later) of learning process for a temporal view.
\textit{(b)} 
See \textbf{Phenomenon 1.2} and \textbf{2.2} for conclusions.
}
\vspace{-0.3cm}
\label{figure:svd_info_ana}
\end{figure*}

\subsection{Temporal SVD Analysis of Policy Learning Path}
\label{subsec:temporal_svd_ana}

The two phenomena observed above naturally raise further questions:
1) The asymmetry of parameter change shown by Phenomenon 1.1 indicates the imbalanced importance of parameters. Can we \textit{rule the important parameters or parameter directions off from the less important ones}?
2) For the detour shown by Phenomenon 2.1, can we \textit{identify the difference among parameters or parameter directions in their detour behaviors}? And do important parameters or parameter directions detour less?


In this paper, we make use a \textit{temporal} viewpoint on policy path.
We introduce Temporal Singular Value Decomposition (SVD) as our main tool for the following investigations.
Given a policy path $\{\theta_1, \theta_2, \dots, \theta_n \}$, we take the parameters of each policy as a parameter vector, i.e., $\theta_i = [ \theta_{i,1}, \theta_{i,2}, \dots, \theta_{n,m} ]$ where $m$ ($\ge 10^5$ usually) is the dimenionality of policy parameters.
We then stack the parameter vectors along the policy path and form a temporal policy parameter matrix, for which standard SVD can be performed:
\begin{equation*}
\begin{aligned}
        \begin{bmatrix}
        \theta_1 \\
        \theta_2 \\
        \vdots \\
        \theta_n
    \end{bmatrix}
    = &
    \begin{bmatrix}
        \theta_{1,1} & \dots & \theta_{1,m} \\
        \theta_{2,1} & \dots & \theta_{2,m} \\
        \vdots & \ddots & \vdots \\
        \theta_{n,1} & \dots & \theta_{n,m} \\
    \end{bmatrix} \\
    = 
    \underbrace{
    \begin{bmatrix}
        u_{1,1} & \dots & u_{1,d} \\
        \vdots & \ddots & \vdots \\
        u_{n,1} & \dots & u_{n,d} \\
    \end{bmatrix}}_{U}
    & \underbrace{
    \begin{bmatrix}
        \sigma_{1} & \dots & 0 \\
        \vdots & \ddots & \vdots \\
        0 & \dots & \sigma_{d} \\
    \end{bmatrix}}_{\Sigma}
    \underbrace{
    \begin{bmatrix}
        v_{1,1} & \dots & v_{1,m} \\
        \vdots & \ddots & \vdots \\
        v_{d,1} & \dots & v_{d,m} \\
    \end{bmatrix}}_{V^{\top}}
\end{aligned}
\label{eq:temporal_svd}
\end{equation*}
where $U, V$ are left and right unitary matrices, the singular values are indexed in a decreasing order with $d = \min(n,m)$.
For the convenience of expression,
we use $u_{i,*}$ for the $i$-th row vector of $U$ and sue $u_{*,j}$ for the $j$-th column vector (and the same way for $V^{\top}$).
The temporal SVD
offers us \textbf{an angle to view how policy evolves in a lower-dimensional space}:
we can now take the row vectors of $U$ as \textit{new $d$-dimensional coordinates} for policies along the policy path (i.e., $u_{i,*}$ for the $i$-th policy), regarding the scaling vector $\texttt{Diag}(\Sigma)$ and the parameter subspace spanned by the row vectors of $V^{\top}$, i.e., $\texttt{Span}(\{ v_{1,*}^{\top}, v_{2,*}^{\top}, \dots, v_{d,*}^{\top} \}) \subseteqq \mathbb{R}^{d}$.
In the following, we also call $\{v_{i,*}\}$ as \textit{SVD directions}.

To answer the two questions listed in the beginning of this subsection,
we introduce two more quantities:
\begin{itemize}
    \item Singular Value Information Amount for a dimensionality number $k \in \{1, \dots, d \}$: $a_k =  \frac{\sum_{i=1}^{k} \sigma_i}{\sum_{i=1}^{d} \sigma_i}$.
    \item SVD Major Dimensionality regarding an information threshold $\beta \in (0,1]$:
    $\mathbb{D}(\beta) = \min \{k \in \{1, \dots, d \}: \alpha_k \ge \beta \}$.
\end{itemize}
Note that $d = \mathbb{D}(1)$ and $\mathbb{D}(0.99)$ recovers \textit{approximate rank} used in a few recent works~\cite{YangZXK20Harnessing,KumarAGL21ImplicitUnder,LyleRD22Understanding}.

We also uses the policy path data obtained by TD3 on MuJoCo and RAD on DMC as in Sec.~\ref{subsec:param_change}.
First, we plot the curves of $\mathbb{D}(\beta)$ against threshold candidates $\beta \in \{0.5, 0.6, 0.7, 0.8, 0.85, 0.9, 0.95, 0.99\}$.
In addition to a layer-wised manner, we consider three periods (i.e., early, middle, later) of learning process for a temporal view.
The results for DMC/walker-walk are shown in Fig.~\ref{subfig:svd_curves_dmc_walker_walk}.
The results are similar in almost all other MuJoCo and DMC environments and the complete figures can be found in Appendix~\ref{appsec:additional_results_svd_info}.

Now we summarize our third phenomenon according to the results in Fig.~\ref{subfig:svd_curves_dmc_walker_walk}.

\begin{mybox}[detach title,before upper={\tcbtitle\quad}]{Phenomenon 1.2 (\textit{Singular Value Information Concentration}):}
The singular value information is highly concentrated on the first a few singular values. The concentration is more evident for the second layer and later periods.
\end{mybox}
This indicates that the policy learning path proceeds mainly in a low-dimensional subspace of the entire parameter space, specially, in the space spanned by the first a few row vectors of $V^{\top}$.
This adds some more explanations to Phenomenon 1.1, telling which parameters change more significantly.
The higher concentration of the second layer also matches the results in Fig.~\ref{figure:ana_param_change},
as the second layer is often more over-parameterized than the other layers.
The concentration of later periods is easy to understand since policy improvement becomes slower gradually.

Further, regarding the second question raised in the beginning,
we expect to find some correlation between the directions of policy parameter evolvement in the low-dimensional subspace obtained by SVD and the improvement of policy performance.
To this end, we study the left SVD unitary matrix, where the row vectors work as new $d$-dimensional coordinates of policies as mentioned above.
In Fig.~\ref{subfig:svd_u_ana_dmc_walker_walk}, we investigate how the policy path evolves along each component of the new coordinate, including plotting the column vectors $\{u_{*,k}\}$ as curves (the second and the third plots), the detour ratio (the fourth plot) and the final change (the fifth plot) for each component.
These results are against the policy performance curve (the first plot of Fig.~\ref{subfig:svd_u_ana_dmc_walker_walk}), and we focus on only the period with significant policy improvement, i.e., the left of the blue dashed vertical line.

According to the results in Fig.~\ref{subfig:svd_u_ana_dmc_walker_walk}, we observe the final phenomenon.

\begin{mybox}[detach title,before upper={\tcbtitle\quad}]{Phenomenon 2.2 (\textit{Policy Evolvement in Major and Minor SVD Directions}):}
The detour of policy update is overall increasingly severe on later coordinate components (corresponding to lower singular values).
Specially, $u_{*,1}$ (the first coordinate component) is almost monotonic while $u_{*,j}$ (the other components) shows harmonic-wave-like changes with overall increasingly higher frequencies for $j \ge 2$.
\end{mybox}
It is surprising to observe Phenomenon 2.2.
Intuitively, it indicates that policy parameter path proceeds monotonically along one major direction while oscillates in other directions with frequencies inversely proportional to singular values.
This empirically supports our hypothesis that important (i.e., major) parameter directions detour less while minor directions detour severely.
The phenomenon also encourages the emergence of theoretical explanations to the evolvement behaviors of policy path in different directions, especially to the harmonic-wave-like changes.
Since typical policy gradients are derived regarding approximate value estimates, we suggest that the dynamics of policy parameters may be closely related to recent studies on the learning dynamics of value function~\cite{LyleROD21,LyleRD22Understanding}.
We leave it as a major direction of future work.

For a brief summary, till now we have observed that policy parameter path evolves mainly in a low-dimensional parameter subspace (spanned by $\{v_{i,*}\}$).
This is commonly seen in our empirical investigations for typical DRL algorithms in popular environments with proprioceptive or visual observations.
A natural idea is:
why not let the agent focus on the policy update in the parameter subspace, by \textit{following the major parameter directions and neglecting the minor directions}? 
Moreover, somewhat excitingly, it seems that $u_{*,1}, u_{*,2}$ has a strong correlation to policy performance.
We are curious about whether it is possible to \textit{leverage the correlation to boost the learning process}.
In the next section, we study on these points along with the proposal of Policy Path Trimming.







\begin{figure*}
\begin{center}
\includegraphics[width=0.9\textwidth]{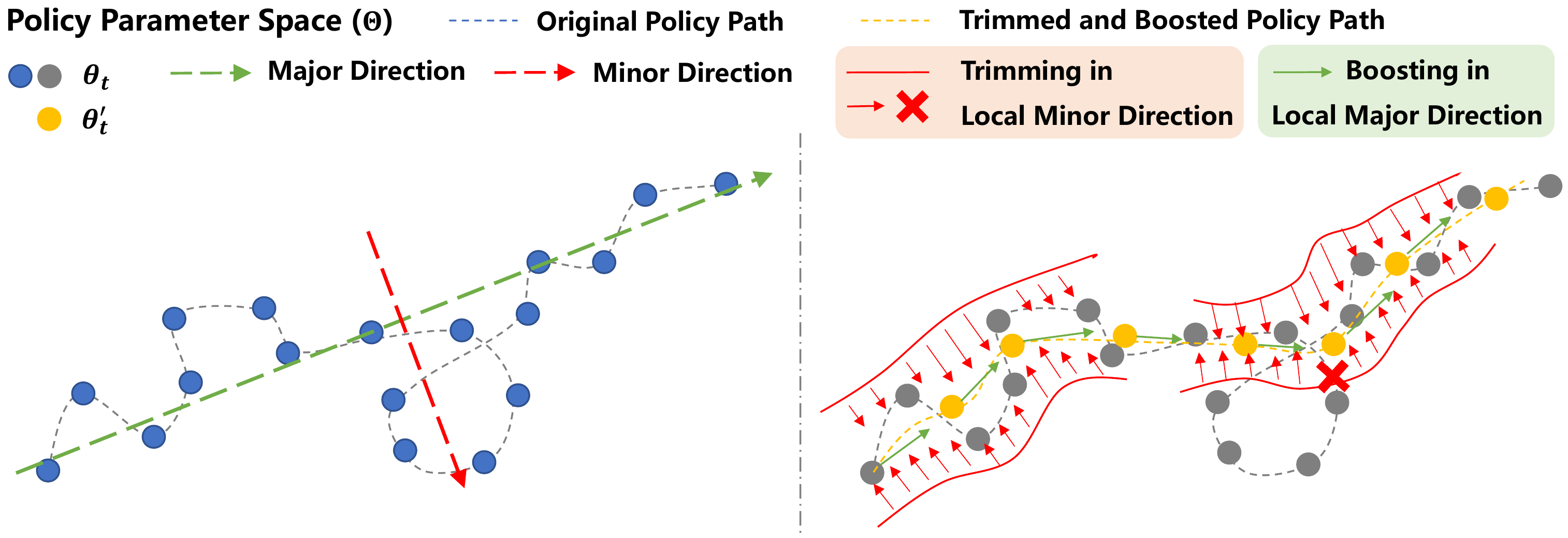}
\end{center}
\caption{A conceptual illustration of Policy Path Trimming and Boosting (PPTB).
The 2D view of an exemplary policy learning path in policy parameter space \textit{(left)} and the corresponding policy learning path improved by PPTB \textit{(right)}.
}
\label{figure:pptb_concept}
\end{figure*}

\section{Policy Path Trimming and Boosting}
\label{sec:ppt_method}

Driven by the phenomena we discovered in the previous section,
we propose a simple and effective method, called Policy Path Trimming and Boosting (\textbf{PPTB}), as a general plug-in improvement to DRL algorithms.
In the following, we introduce the details of the two components of PPTB, i.e., 
Policy Path Trimming (Sec.~\ref{subsec:ppt}) and Policy Path Boosting (Sec.~\ref{subsec:ppb}),
and then the general implementation of DRL with PPTB (Sec.~\ref{subsec:drl_with_pptb}).

\subsection{Policy Path Trimming}
\label{subsec:ppt}

As summarized in Phenomenon 1.2, we have observed that the policy path mainly evolves in a low-dimensional parameter subspace with a large proportion of singular value information concentrated in the first a few singular values.
Our first idea is to truncate the parameter change in minor SVD directions and only remain the change in major ones based on Temporal SVD of policy path.
We call this method as Policy Path Trimming (PPT).

Given a policy path $\{\theta_1, \theta_2, \dots, \theta_n \}$ and the number of major directions to remain $r_t$ (usually $\ll d = \min(n,m)$). 
For a policy with original parameters $\theta_i$,
PPT reconstructs policy parameters by only taking into consideration of the first $r_t$ SVD directions:
\begin{equation}
    \tilde{\theta}_i = 
    \underbrace{
    \begin{bmatrix}
        u_{i,1} & \dots & u_{i,\textcolor{blue}{r_t}} \\
    \end{bmatrix}}_{u_{i,1:\textcolor{blue}{r_t}}}
    \underbrace{
    \begin{bmatrix}
        \sigma_{1} & \dots & 0 \\
        \vdots & \ddots & \vdots \\
        0 & \dots & \sigma_{\textcolor{blue}{r_t}} \\
    \end{bmatrix}}_{\Sigma[ 1:\textcolor{blue}{r_t} ]}
    \underbrace{
    \begin{bmatrix}
        v_{1,*}\\
        \vdots \\
        v_{\textcolor{blue}{r_t},*} \\
    \end{bmatrix}}_{V^{\top}[ 1:\textcolor{blue}{r_t} ]}
\label{eq:ppt}
\end{equation}
Note that we use $\cdot,\cdot$ in subscripts and $[\cdot,\cdot]$ to denote the slices of vector and matrix respectively.

A conceptual illustration of PPT is shown in Fig.~\ref{figure:pptb_concept}.
As shown by the red arrows and cross, PPT trims the policy parameter update in minor SVD directions and enforces the policy path proceeds in major directions, i.e., in the subspace $\texttt{Span}(\{ v_{1,*}^{\top}, v_{2,*}^{\top}, \dots, v_{r_t,*}^{\top} \})$.
Intuitively, this suppresses the detours and oscillations of parameter update (recall the results in Fig.~\ref{subfig:svd_u_ana_dmc_walker_walk}).
Therefore, we expect PPT to improve the efficiency of policy learning process in this sense.
One may worry about whether the trimmed policy parameters $\theta_i^{\prime}$ still ensure an effective policy.

For sanity check, we compare policy performance between $\theta_i$ and $\theta_i^{\prime}$ regarding different choices of $r_t$
in Appendix~\ref{appsec:additional_results_svd_recon}.
We found that a small $r_t$ is sufficient to ensure a valid recovery of policy performance, while increasing $r_t$ shows no significant difference.

\subsection{Policy Path Boosting}
\label{subsec:ppb}

In addition to suppressing the parameter change in minor directions,
we are interested in accelerating policy learning by leveraging the correlation between policy performance and major SVD directions as noted in Phenomenon 2.2.
In particular,
we propose to boosting the change of $u_{*,1}, u_{*,2}$ since they show near monotonic changes corresponding to the improvement of policy performance in Fig.~\ref{subfig:svd_u_ana_dmc_walker_walk}.
We call this method as Policy Path Boosting (PPB).

For a policy with original parameters $\theta_i$ among the policy path $\{\theta_1, \theta_2, \dots, \theta_n \}$,
PPB modifies $\theta_i$ by increasing the value of $u_{i,1}, u_{i,2}$ along the temporal direction with the amplitude $p_b$:
\begin{equation}
\begin{aligned}
    \textcolor{red}{\hat{u}_{i,*}} & = u_{i,*} + p_b (u_{n,*} - u_{1,*}) \\
    \hat{\theta}_i & = 
    \underbrace{
    \begin{bmatrix}
        \textcolor{red}{\hat{u}_{i,1}} & \textcolor{red}{\hat{u}_{i,2}} & u_{i,3} & \dots & u_{i,d} \\
    \end{bmatrix}}_{\text{concat}(\textcolor{red}{\hat{u}_{i,1:2}} \ , \ u_{i,3:d})}
    \Sigma
    V^{\top}
\end{aligned}
\label{eq:ppb}
\end{equation}
PPB only modifies $u_{i,1}, u_{i,2}$ while keeps other parts unchanged.
As illustrated by the green arrows in Fig.~\ref{figure:pptb_concept},
PPB boosts policy parameter update in the major direction to accelerate the improvement of policy performance.

\begin{algorithm}[t]
\small
\SetAlgoLined
    \PyComment{env: environment} \\
    \PyComment{agent: policy-based DRL agent} \\
    \PyComment{d\_p: policy parameter buffer with size $k$} \\
    \PyComment{$r_{t},r_{b}, p_b$: dimensionality hyperparams of PPTB where $r_{t} \ge r_{b}$ (see Eq.~\ref{eq:ppt} and~\ref{eq:ppb})} \\
    \PyComment{$t_{s},t_{p}$: intervals of storing policy and performing PPTB where $t_{p}$ \% $t_{s} == 0$} \\
    \PyCode{} \\
    \PyCode{d\_p = [agent.policy.params] } \PyComment{Initialize} \\
    \PyCode{for $t$ in range(max\_interaction\_steps):} \\
        \QuadPyComment{Typical agent-env interaction (omitted)} \\
        \PyCode{\quad ...} \\
        \PyCode{\quad agent.learn()} \\
        \PyCode{} \\
        \QuadPyComment{1) Store policy parameters at intervals} \\
        \PyCode{\quad If $t$ \%  $t_s == 0$} \\
            \PyCode{\quad \quad d\_p.append(agent.policy.params)} \\
            \PyCode{\quad \quad d\_p = d\_p[$-k$:]} \\
        \PyCode{} \\
        \QuadPyComment{2) Perform Temporal SVD and PPTB} \\
        \PyCode{\quad If $t$ \%  $t_p == 0$} \\
            \PyCode{\quad \quad u, sgl, vh = SVD(d\_p)} \\
            \PyCode{\quad \quad u\_b = (u[$-1$] - u[$0$]}) * $p_b$ + u[$-1$] \\
            \PyCode{\quad \quad u\_tb = concat(u\_b[:$r_b$], u[$-1$,$r_b$:$r_t$])} \\
            \PyCode{\quad \quad param\_tb = (u\_tb * sgl[:$r_t$]}).dot(vh[:$r_t$]) \\
        \PyCode{} \\
        \QuadPyComment{3) Apply modified params to agent} \\
            \PyCode{\quad \quad agent.policy.load(params\_tb)}
\caption{DRL with PPTB (Psedudocode in a PyTorch-like style)}
\label{algo:pptb}
\end{algorithm}

\subsection{DRL with PPTB}
\label{subsec:drl_with_pptb}

Now, we are ready to propose PPTB as a combination of PPT and PPB.
Formally,
\begin{equation}
\label{eq:pptb}
    \theta_i^{\prime} = 
    \text{concat}(\textcolor{red}{\hat{u}_{i,1:2}} \ , \ u_{i,3:\textcolor{blue}{r_t}})
    \Sigma[1:\textcolor{blue}{r_t}]
    V^{\top}[1:\textcolor{blue}{r_t}].
\end{equation}
Although we can perform PPTB for any policy on an arbitrary policy path,
in practice we consider the policy path that consists of the historical policies within a recent window and the current policy, and we perform PPTB for current policy.

Apparently, PPTB is algorithmic-agnostic.
For almost all off-the-shelf policy-based DRL algorithms,
PPTB can be implemented and incorporated in by adding the following three steps to conventional policy update scheme:
\begin{itemize}
    \item[1)] Initialize a policy buffer and store the parameters of current policy at intervals along the policy learning process.
    \item[2)] At certain occasions, retrieve the recent policy parameter path and perform Temporal SVD; then do policy path trimming and boosting for current policy.
    \item[3)] Load the consequent policy parameters processed by PPTB back to current policy.
\end{itemize}
We provide a pytorch-like pseudocode of PPTB implementation in Algorithm~\ref{algo:pptb}, where the slices of vector and matrix are also in a pytorch-like style.
In our practical implementation used by our experiments, we only make the modifications of about 10-line core codes in the official implementations of TD3~\cite{Fujimoto2018TD3} and RAD~\cite{LaskinLSPAS20RAD}.
Note that the main computation cost added by PPBT is the calculation of SVD, which can be expensive when policy parameter dimensionality $m$ and buffer size $k$ are large.
In practice, we perform PPBT at sparse intervals 
(e.g., $t_p \ge 1000$) and the time cost becomes acceptable.

For typical value-based DRL algorithms, we can also treat the value network as a special form of policy and perform PPTB in the same way.
However, this is kind of reckless since the natures of policy and value function are different, nor are the learning dynamics of them.
A recent phenomena called \textit{Policy Churn} discovered in~\cite{Schaul22PolicyChurn} reveals that the greedy policy induced by a typical value network changes its actions on about 10\% states after one update.
We leave in-depth studies on value function path in the future.

\begin{table}[t]
  \caption{Empirical results of Policy Path Trimming and Boosting (PPTB) for TD3~\cite{Fujimoto2018TD3} in four MuJoCo environments. Means and standard deviation errors across six independent trials are reported. The improvements ($\nearrow$) and the aggregate results are calculated according to a \textit{random-agent} baseline (whose scores are omitted). 
  }
  \label{table:mujoco_overall_evaluations}
\begin{center}
\begin{small}
\begin{sc}
\scalebox{0.9}{
\begin{tabular}{c|c|c|c}
\toprule 
Environments & Metrics & TD3 & TD3-PPTB \\
\midrule
\multirow{2}*{HalfCheetah} & Score & 10548 $\pm$ 357 & 1110 $\pm$ 60 (\textcolor{blue}{5.12\%$\nearrow$}) \\
\cmidrule(r){2-4}
 & \colorbox{gray!20}{AUC} & 7981 $\pm$ 304 & 8689 $\pm$ 76 (\textcolor{blue}{8.55\%$\nearrow$}) \\
\midrule
\multirow{2}*{Hopper} & Score & 3394 $\pm$ 59 & 3420 $\pm$ 54 (\textcolor{blue}{0.77\%$\nearrow$}) \\
\cmidrule(r){2-4}
 & \colorbox{gray!20}{AUC} & 2005 $\pm$ 99 & 2249 $\pm$ 114 (\textcolor{blue}{12.28\%$\nearrow$}) \\
\midrule
\multirow{2}*{Walker2d} & Score & 3406 $\pm$ 436 & 4385 $\pm$ 164 (\textcolor{blue}{28.76\%$\nearrow$}) \\
\cmidrule(r){2-4}
 & \colorbox{gray!20}{AUC} & 1977 $\pm$ 353 & 2805 $\pm$ 85 (\textcolor{blue}{41.92\%$\nearrow$}) \\
\midrule
\multirow{2}*{Ant} & Score & 4177 $\pm$ 451 & 5147 $\pm$ 449 (\textcolor{blue}{22.81\%$\nearrow$}) \\
\cmidrule(r){2-4}
 & \colorbox{gray!20}{AUC} & 2624 $\pm$ 293 & 3542 $\pm$ 392 (\textcolor{blue}{34.02\%$\nearrow$}) \\
\midrule
\multirow{2}*{Aggregate} & Score & 1.0 &  1.1436  \\
\cmidrule(r){2-4}
 & \colorbox{gray!20}{AUC} & 1.0  &  1.2419  \\
\bottomrule
\end{tabular}
}
\end{sc}
\end{small}
\end{center}
\vskip -0.1in
\end{table}

\section{Experimental Evaluation of PPTB}
\label{sec:exps}


\paragraph{Setups}

To evaluate the performance of our proposed method PPTB,
we consider the continuous control environments in OpenAI MuJoCo~\cite{Brockman2016Gym} and DeepMind Control Suite (DMC)~\cite{Tassa2018DMC}, including both proprioceptive and visual inputs.
Concretely, we use TD3~\cite{Fujimoto2018TD3} and RAD~\cite{LaskinLSPAS20RAD} as the base algorithms for MuJoCo and DMC respectively, thanks to their simplicity and effectiveness.
We use the official codes of TD3 and RAD and modify them according to Algorithm 1 to implement PPTB with no other change to the original implementation.

Along with the implementation of PPTB on TD3 and RAD, three additional hyperparameters need to be considered.
For PPT, we mainly consider to choose the number of major directions to remain (i.e., $r_t$ in Sec.~\ref{subsec:ppt}) in the set $\{8,16,32,64,128\}$ for each environment.
For the maintenance of the policy path, we save the policy parameters every 25 mini-batch gradient updates of the policy network in a FIFO policy buffer with a size of 2k and 1k for MuJoCo and DMC respectively.
Note that for PPB, we boost the first two major SVD directions (i.e., $u_{*,1}, u_{*,2}$ as described in Sec.~\ref{subsec:ppb}) currently.
The number here may not be optimal and can be different specially in other environments.

We train the agent for 1 million time steps and evaluate the agent every 5k time steps for each agent-environment configuration.
We run each configuration with six random seeds.
We consider two evaluation metrics:
1) SCORE: the \textit{maximum} of average (over multiple runs) evaluation returns over the course of learning, which is used by TD3~\cite{Fujimoto2018TD3};
2) AUC: the \textit{mean} of average evaluation returns over the course of learning, which is also used in~\cite{KumarA0CTL22DR3}.
The former cares about \textit{effectiveness} while the later measures \textit{efficiency} and \textit{stability}, which is also significant to practical use of DRL algorithms.

For MuJoCo environments, we report SCORE and AUC for 1M time step training; for DMC, we report for 100k, 500k (and 1M for cheetah-run) as usually done in prior works~\cite{LaskinLSPAS20RAD}.
For the convenience of comparing across different return scales, we also report the normalized results with a random-agent baseline as 0 and the DRL base algorithm (i.e., TD3 or RAD) as 1.

\paragraph{Results}

The results for MuJoCo and DMC are reported in Table~\ref{table:mujoco_overall_evaluations} and~\ref{table:dmc_overall_evaluations}, respectively.
The results show the overall improvements brought by PPTB for both TD3 and RAD, indicating its compatibility and effectiveness.
The improvement is more obvious in AUC, which means PPTB improves the learning efficiency and stability of the base algorithms.
Somewhat surprisingly,
we can observe that PPTB outperforms the base algorithms by a large margin in several environments, e.g., Walker2d and Ant in MuJoCo, and Walker-walk in DMC.
To some extent, this reveals the potential of studying and modifying the policy learning path in achieving general improvements to DRL agents.





\begin{table}[t]
  \caption{Empirical results of Policy Path Trimming and Boosting (PPTB) for RAD~\cite{LaskinLSPAS20RAD} in four DeepMind Control (DMC) environments.
  Means and standard deviation errors across six independent trials are reported. The improvements ($\nearrow$) and the aggregate results are calculated according to a \textit{random-agent} baseline (whose scores are omitted). 
  }
  \label{table:dmc_overall_evaluations}
\begin{center}
\begin{small}
\begin{sc}
  \scalebox{0.9}{
  \begin{tabular}{c|c|c|c}
    \toprule 
    Environments & Metrics & RAD & RAD-PPTB \\
    \midrule
    \multirow{4}*{finger-spin} & Score (100k) & 553 $\pm$ 71 & 592 
$\pm$ 55 (\textcolor{blue}{7.09\%$\nearrow$}) \\
    \cmidrule(r){2-4}
     & \colorbox{gray!20}{AUC (100k)} & 201 $\pm$ 33 &  254 $\pm$ 40 (\textcolor{blue}{26.76\%$\nearrow$}) \\
    \cmidrule(r){2-4}
     & Score (500k) & 898 $\pm$ 55 & 970 $\pm$ 5 (\textcolor{blue}{8.04\%$\nearrow$}) \\
    \cmidrule(r){2-4}
     & \colorbox{gray!20}{AUC 500k} & 695 $\pm$ 41 & 767 $\pm$ 20 (\textcolor{blue}{10.40\%$\nearrow$}) \\
    \midrule
    
    \multirow{4}*{walker-walk} & Score (100k) & 210 $\pm$ 48 & 349 
$\pm$ 35 (\textcolor{blue}{86.87\%$\nearrow$}) \\
    \cmidrule(r){2-4}
     & \colorbox{gray!20}{AUC 100k} & 104 $\pm$ 22 &  148 $\pm$ 15 (\textcolor{blue}{81.48\%$\nearrow$}) \\
    \cmidrule(r){2-4}
     & Score (500k) & 923 $\pm$ 9 & 940 
$\pm$ 6 (\textcolor{blue}{1.94\%$\nearrow$}) \\
    \cmidrule(r){2-4}
     & \colorbox{gray!20}{AUC 500k} & 583 $\pm$ 30 & 672 $\pm$ 11 (\textcolor{blue}{16.69\%$\nearrow$}) \\
    \midrule
    
    \multirow{4}*{cartpole-swingup} & Score (100k) & 235 $\pm$ 15 & 268 
$\pm$ 18 (\textcolor{blue}{30.84\%$\nearrow$}) \\
    \cmidrule(r){2-4}
     & \colorbox{gray!20}{AUC 100k} & 158 $\pm$ 6 &  189 $\pm$ 2 (\textcolor{blue}{103.33\%$\nearrow$}) \\
    \cmidrule(r){2-4}
     & Score (500k) & 836 $\pm$ 12 & 860 $\pm$ 9 (\textcolor{blue}{3.38\%$\nearrow$}) \\
    \cmidrule(r){2-4}
     & \colorbox{gray!20}{AUC 500k} & 530 $\pm$ 18 & 572 $\pm$ 20 (\textcolor{blue}{10.44\%$\nearrow$}) \\
    \midrule
    
    \multirow{4}*{cheetah-run} & Score (100k) & 360 $\pm$ 8 & 394 
$\pm$ 10 (\textcolor{blue}{10.11\%$\nearrow$}) \\
    \cmidrule(r){2-4}
     & \colorbox{gray!20}{AUC 100k} & 194 $\pm$ 5 &  207 $\pm$ 13 (\textcolor{blue}{7.64\%$\nearrow$}) \\
    \cmidrule(r){2-4}
     & Score (500k) & 574 $\pm$ 13 & 605 $\pm$ 10 (\textcolor{blue}{5.63\%$\nearrow$}) \\
    \cmidrule(r){2-4}
     & \colorbox{gray!20}{AUC 500k} & 428 $\pm$ 10 & 452 $\pm$ 6 (\textcolor{blue}{5.94\%$\nearrow$}) \\
    \cmidrule(r){2-4}
     & Score (1m) & 700 $\pm$ 8 & 730 $\pm$ 10 (\textcolor{blue}{4.43\%$\nearrow$}) \\
    \cmidrule(r){2-4}
     & \colorbox{gray!20}{AUC 1m} & 534 $\pm$ 8 & 554 $\pm$ 4 (\textcolor{blue}{3.92\%$\nearrow$}) \\
    \midrule
    \multirow{4}*{Aggregate} & Score (100k) & 1.0 &  1.3372   \\
    \cmidrule(r){2-4}
     & \colorbox{gray!20}{AUC 100k} & 1.0  &  1.5480   \\
    \cmidrule(r){2-4}
     & Score (500k) & 1.0 &  1.0474  \\
    \cmidrule(r){2-4}
     & \colorbox{gray!20}{AUC 500k} & 1.0  &  1.1086  \\
    \bottomrule
  \end{tabular}
}
\end{sc}
\end{small}
\end{center}
\vskip -0.1in
\end{table}

\section{Related Works}
\label{sec:related_work}

In the topic of general Deep Learning, there have been a large number of efforts devoted consistently during the past decade to understand the learning dynamics and behaviors of deep neural networks, e.g., Neural Tangent Kernel (NTK)~\cite{JacotHG18NTK} and Lazy Training~\cite{Chizat2018OnLazyTraining}.
Recently, there are a few works that study the learning dynamics of DRL agents from different perspectives.
Among them, a major stream of works study the \textit{co-learning dynamics} between representation and RL functions (i.e., value network or policy network)~\cite{DabneyBRDQBS21VIP,KumarAGL21ImplicitUnder,LyleRD22Understanding,NikishinSDBC22PrimacyBias}.
The focus of this stream is that DRL agent over-shapes its representation towards early experiences and objectives (e.g., TD targets of early policies)
while becomes less capable for later learning process.
Such a degradation becomes more severe and detrimental gradually due to the non-stationary learning nature, finally leading to myopic convergence or even divergence.
This is also called as \textit{Implicit Underparameterization}~\cite{KumarAGL21ImplicitUnder}, \textit{Capacity Loss}~\cite{LyleRD22Understanding}, \textit{Primacy Bias}~\cite{NikishinSDBC22PrimacyBias} and etc.
This is usually studied in value function learning since bootstrapping is one major source of the learning issues~\cite{Achiam2019Towards}.
In this work, we study the learning dynamics of typical DRL policy networks, mainly from the angle of policy network parameters.
To the best of our knowledge, we are almost the first to study how the parameters of practical DRL policy networks evolve.

Recently, a few simple yet effective methods are proposed as generic improvements to DRL algorithms,
e.g., ITER~\cite{IglFLBW21Iter}, InFeR~\cite{LyleRD22Understanding},
Periodically Reset~\cite{NikishinSDBC22PrimacyBias},
DR3~\cite{KumarA0CTL22DR3}.
Fourier Feature Network~\cite{YangAA22FFN},
Spectral Normalization~\cite{GogianuBRCBP21SpectralNorm}.
A concurrent work~\cite{Sokar2302Dormant} presents the Dormant Neuron Phenomenon in DRL: the neurons of typical value networks gradually become inactive during the learning process, leading to the loss of expressivity.
This phenomenon aligns well to previous efforts made in studying co-learning dynamics we mentioned above.
A new solution called ReDo is proposed to recycle dormant neurons throughout training.
Different from these works, we propose our method based on the phenomena discovered during our empirical investigation on the learning path of typical policy network.

A related work that also studies in a view of policy learning path is ~\cite{tang2022PeVFA}, where a new extension of value function called Policy-extended Value Function Approximator (PeVFA) is proposed.
Through taking policy representation (embedding) as input, PeVFA is empowered to 
preserve the values of multiple policies
and wield the appealing value generalization among policies to improve generalized policy iteration~\cite{Sutton1988ReinforcementLA}.
In this paper, we focus on in-depth investigations on the dynamics of policy parameters while \cite{tang2022PeVFA} aims at improving value function approximation and generalization.

\section{Limitations}
\label{sec:limitations}

In this paper, we only provide empirical investigations on the phenomena we present in Sec.~\ref{sec:phen}.
Although we observe relatively consistent results for TD3 and RAD across a variety of MuJoCo and DMC environments, which demonstrates the generality to some degree,
we have no theoretical supports for the observed phenomena at present.
A rudimentary thought on this point is to study the learning dynamics of $\nabla_{\theta} J(\pi_{\theta})$, especially of $\nabla_{a} Q_{\phi}(s,a)$.
This is because the policy path is the accumulation of policy gradients while policy gradients significantly determined by the landscape of $Q$-network that learns concurrently along the process. We believe that recent studies on the learning dynamics of value function and representation~\cite{LyleROD21,LyleRD22Understanding} can be inspiring reference.

Still from the empirical perspective, our work is not complete in the sense that we only consider TD3 and RAD (i.e., SAC inside).
On-policy policy-based DRL algorithms like PPO~\cite{Schulman2017PPO} may have a very different policy path.
Besides, the empirical investigation for value-based DRL algorithms like DQN~\cite{Mnih2015DQN} and its variants, i.e., the \textit{value function path}, are expected in the future.

For methodology,
we propose very simple methods and we believe that there is great potential in more sophisticated methods to be proposed.
First, we use standard SVD as the main tool in both our empirical investigation and our methods.
We use no acceleration for SVD nor other more advanced alternatives to obtain the major and minor directions of policy path.
Moreover, we use fixed dimensionalities and intervals for policy path trimming and boosting.
This paper contains no attempts in designing adaptive approaches or proposing principled algorithms (which may rely on the advance in theoretical results).
A another limitation is that our proposed method is not evaluated sparse-reward environments.

\section{Conclusion}
\label{sec:conclusion}


In this paper, we present a few attempts in unveiling the learning dynamics of policy network over the training course of typical DRL agents, mainly from an empirical perspective.
Focusing on the concept of policy learning path, we conduct a series of empirical investigations and summarize four common phenomena, revealing that the policy learning path of typical DRL agents evolves mainly in a low-dimentional space with monotonic and waggling changes in major and minor parameter directions respectively.
Driven by our discovery, we propose a simple method, called Policy Path Trimming and Boosting (PPTB), as a plug-in improvement to general DRL algorithms.
We demonstrate the effectiveness of PPTB
based on TD3 and RAD
in a few MuJoCo and DMC environments.


\bibliography{ppt_arXiv}
\bibliographystyle{plain}

\newpage
\appendix
\onecolumn
\section{Additional Experimental Results}
\label{appsec:additional_results}


\subsection{Empirical Investigation on Policy Parameter Change Amount}
\label{appsec:additional_results_change_amount}

\begin{figure*}[h]
\begin{center}
\subfigure[HalfCheetah-v4]{
\includegraphics[width=1.0\textwidth]{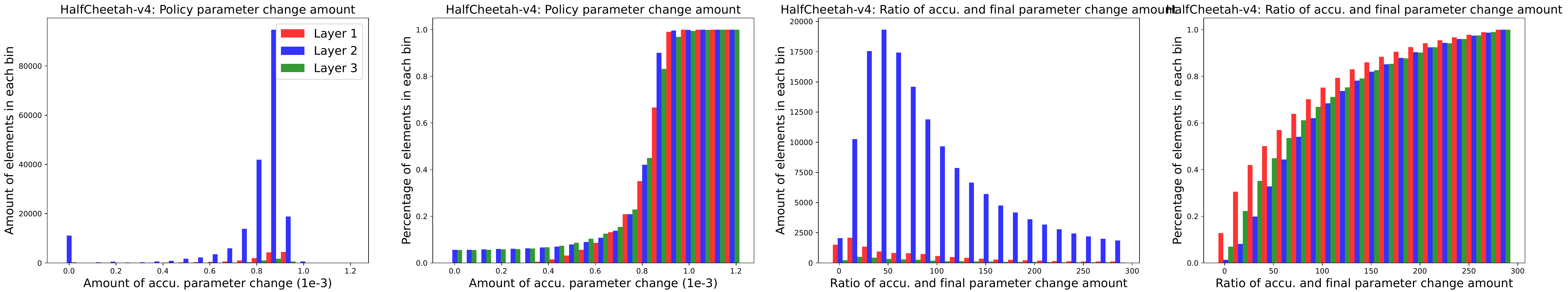}
}
\vspace{-0.2cm}
\subfigure[Hopper-v4]{
\includegraphics[width=1.0\textwidth]{figs/layer_param_change_analysis_Hopper-v4_sn3_20221209.pdf}
}
\vspace{-0.2cm}
\subfigure[Walker2d-v4]{
\includegraphics[width=1.0\textwidth]{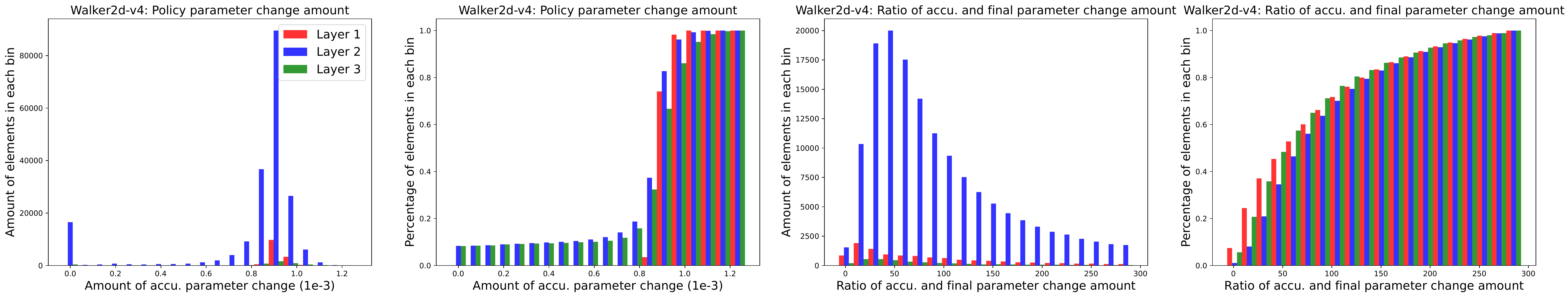}
}
\vspace{-0.2cm}
\subfigure[Ant-v4]{
\includegraphics[width=1.0\textwidth]{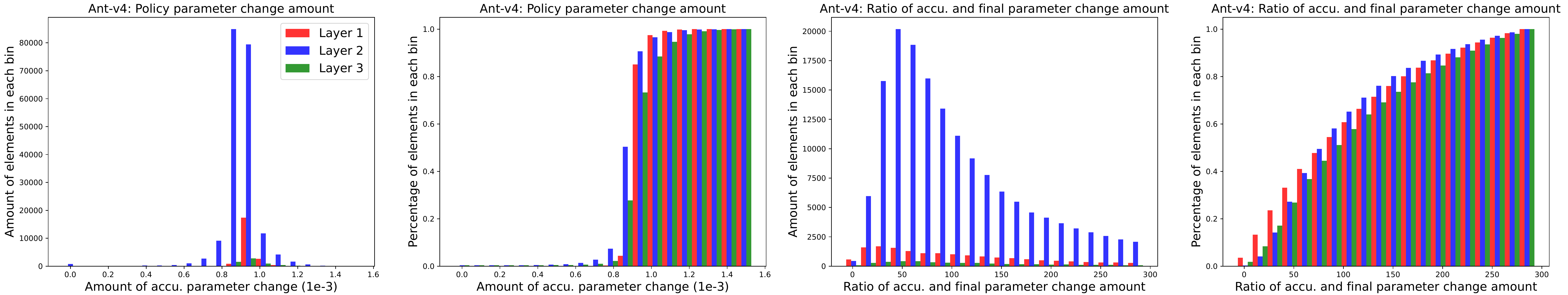}
}
\vspace{-0.2cm}
\subfigure[Humanoid-v4]{
\includegraphics[width=1.0\textwidth]{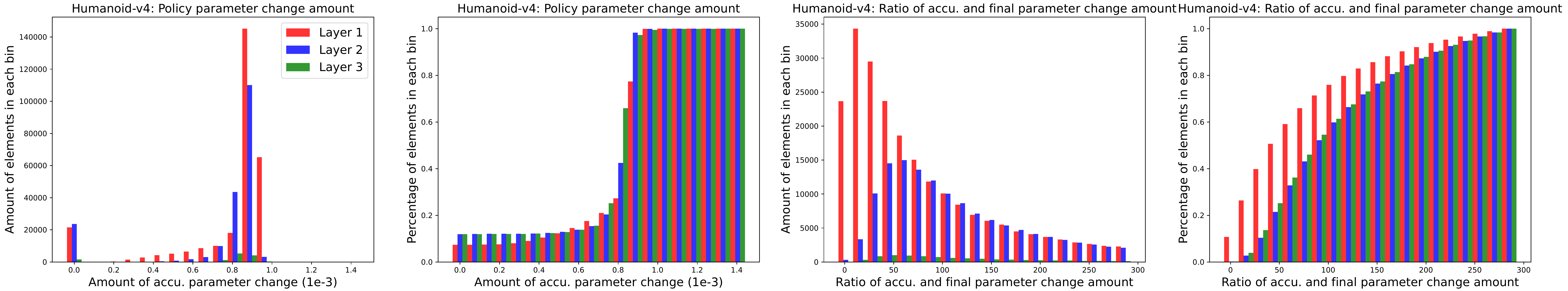}
}
\end{center}
\vspace{-0.3cm}
\caption{Full results of empirical investigation on policy parameter change amount in MuJoCo environments.
}
\label{figure:full_ana_param_change_mujoco}
\end{figure*}

\begin{figure*}
\begin{center}
\vspace{-0.3cm}
\subfigure[cartpole-swingup]{
\includegraphics[width=1.0\textwidth]{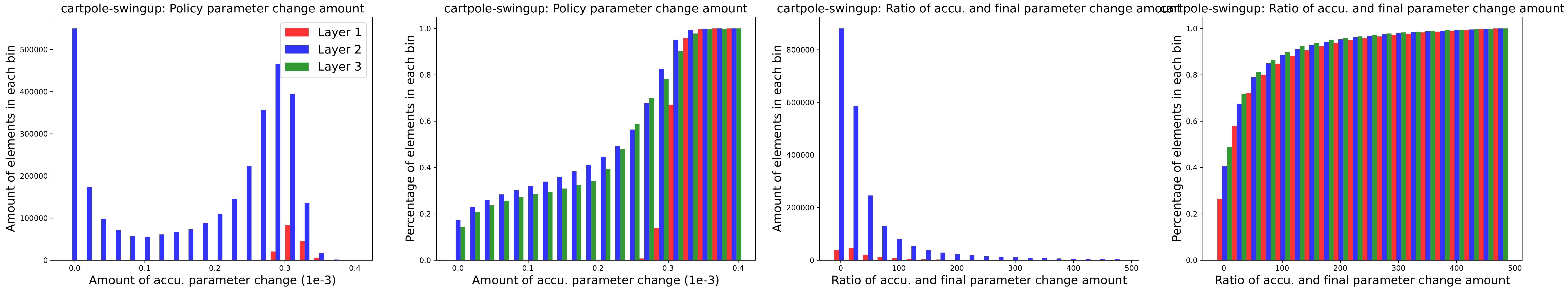}
}
\vspace{-0.2cm}
\subfigure[finger-spin]{
\includegraphics[width=1.0\textwidth]{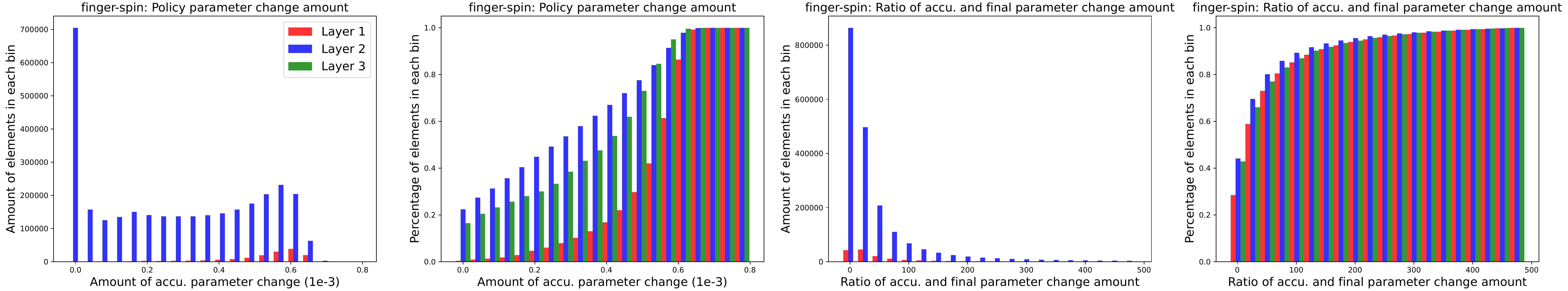}
}
\vspace{-0.2cm}
\subfigure[walker-walk]{
\includegraphics[width=1.0\textwidth]{figs/rad_layer_param_change_analysis_walker-walk_sn3_tr500_20221209.pdf}
}
\vspace{-0.2cm}
\subfigure[hopper-stand]{
\includegraphics[width=1.0\textwidth]{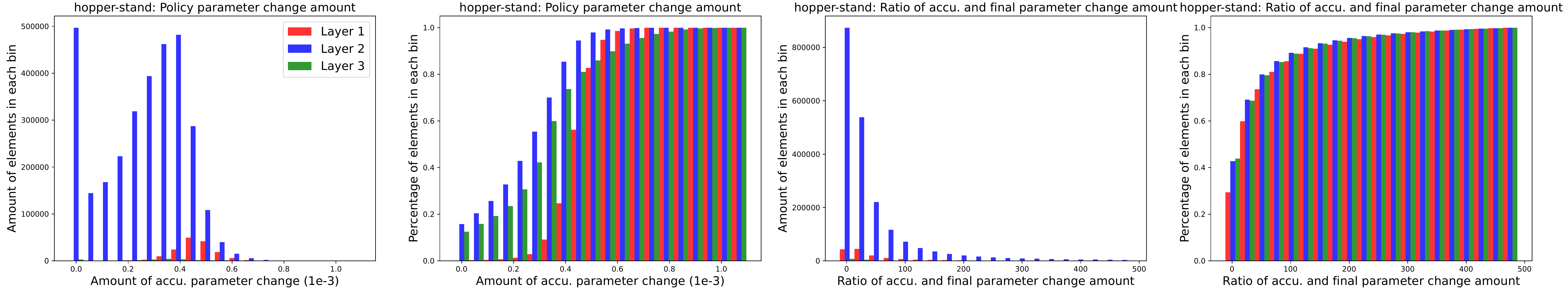}
}
\vspace{-0.2cm}
\subfigure[cheetah-run]{
\includegraphics[width=1.0\textwidth]{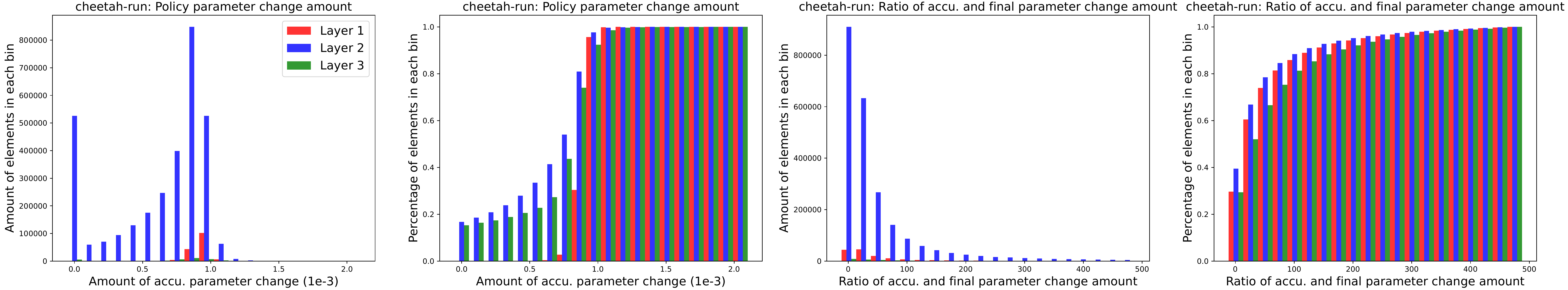}
}
\end{center}
\vspace{-0.3cm}
\caption{Full results of empirical investigation on policy parameter change amount in DMC environments.
}
\label{figure:full_ana_param_change_dmc}
\end{figure*}

\clearpage
\subsection{Empirical Investigation on Policy Learning Path by Temporal SVD}
\label{appsec:additional_results_svd_info}

\begin{figure*}[h]
\begin{center}
\subfigure[cartpole-swingup]{
\includegraphics[width=0.85\textwidth]{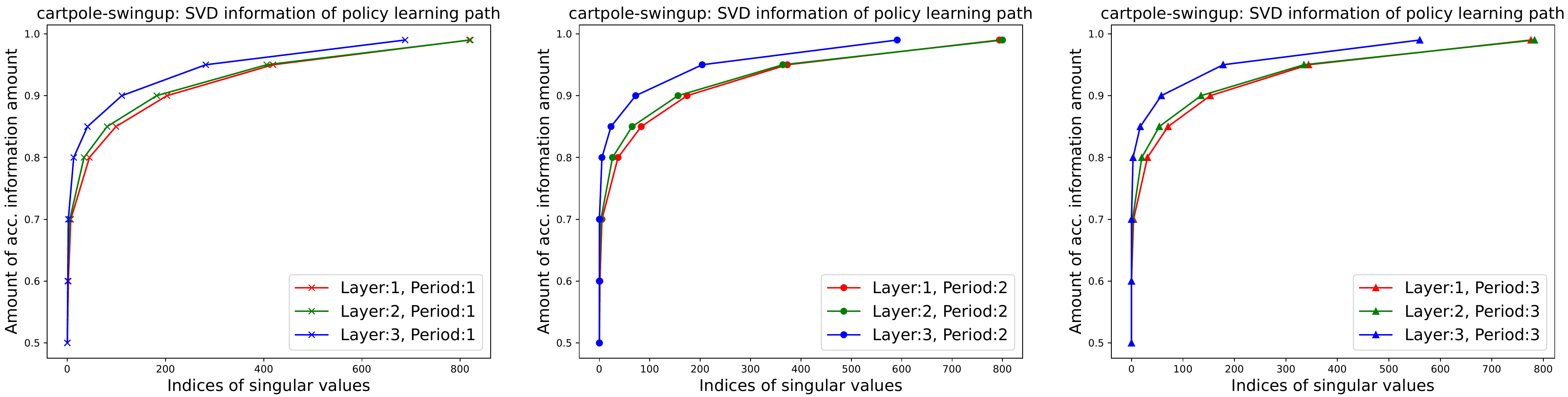}
}
\vspace{-0.2cm}
\subfigure[finger-spin]{
\includegraphics[width=0.85\textwidth]{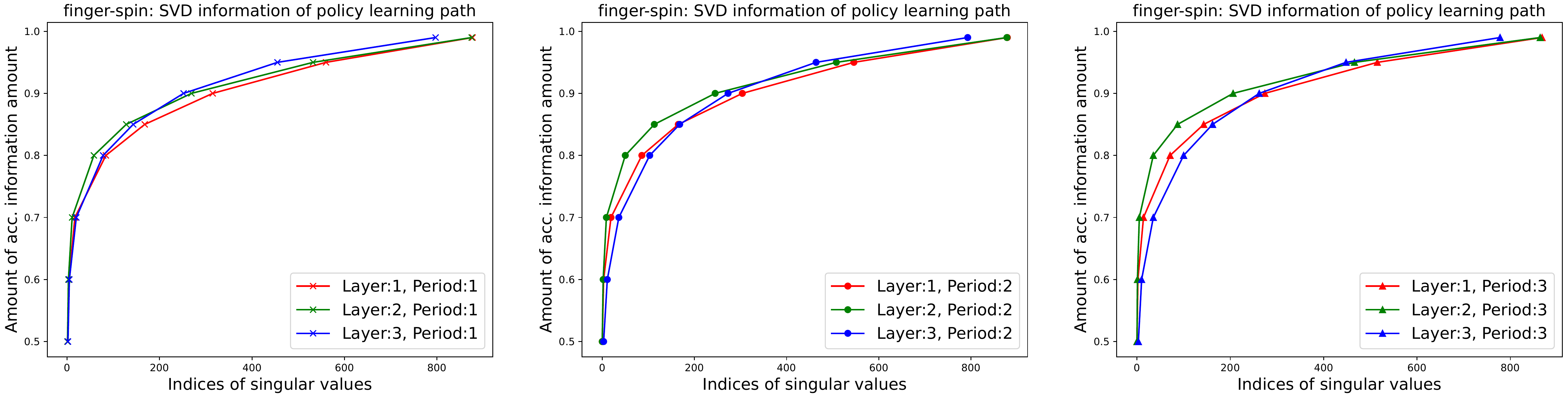}
}
\vspace{-0.2cm}
\subfigure[walker-walk]{
\includegraphics[width=0.85\textwidth]{figs/rad_layer_svd_raw_analysis_walker-walk_sn1_20221214.pdf}
}
\vspace{-0.2cm}
\subfigure[hopper-stand]{
\includegraphics[width=0.85\textwidth]{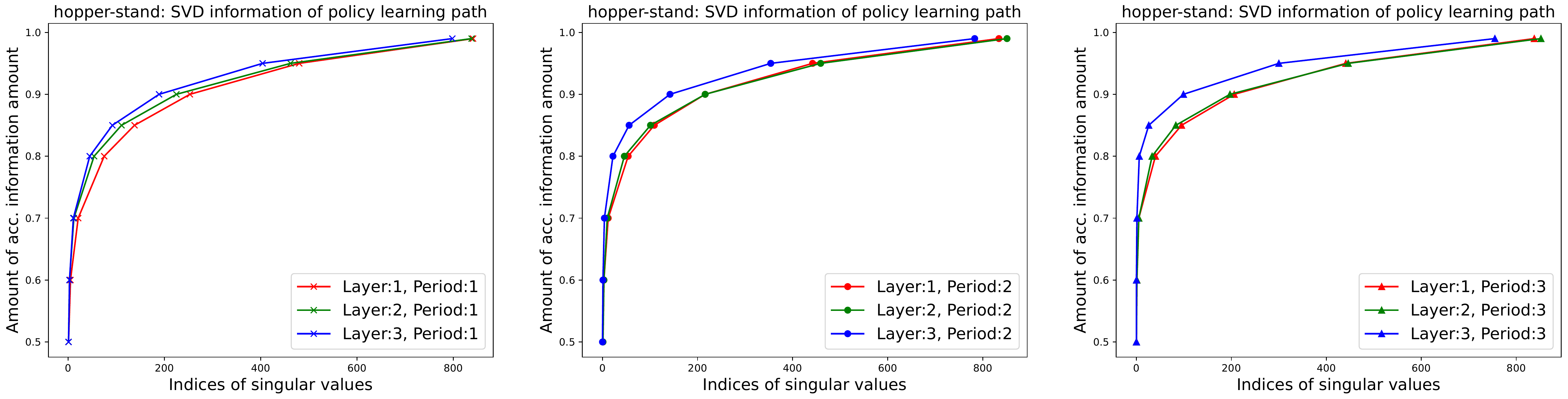}
}
\vspace{-0.2cm}
\subfigure[cheetah-run]{
\includegraphics[width=0.85\textwidth]{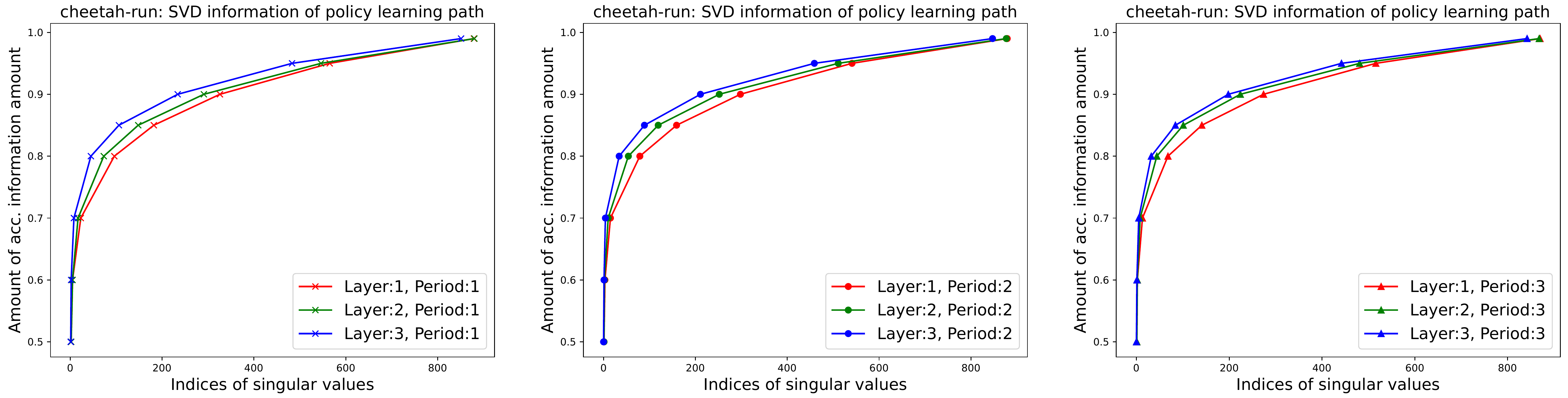}
}
\end{center}
\vspace{-0.3cm}
\caption{Full results of empirical investigation on SVD Information of policy learning path in DMC environments.
}
\label{figure:full_svd_info_dmc}
\end{figure*}

\begin{figure*}
\begin{center}
\vspace{-0.3cm}
\subfigure[HalfCheetah-v4]{
\includegraphics[width=1.0\textwidth]{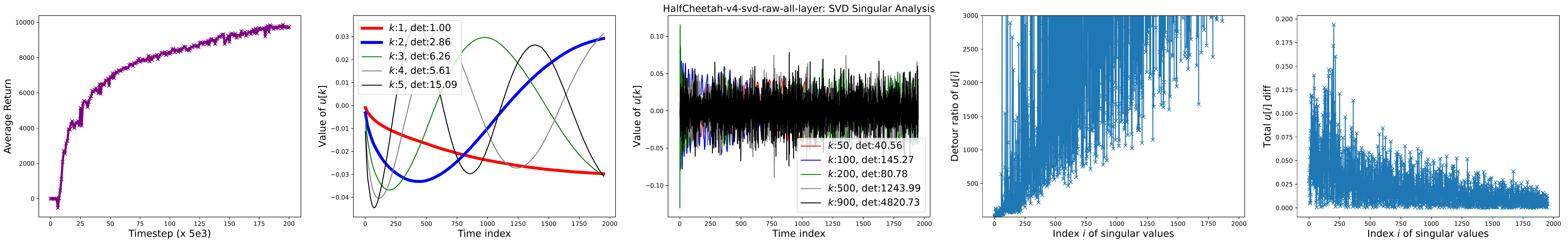}
}
\vspace{-0.2cm}
\subfigure[Hopper-v4]{
\includegraphics[width=1.0\textwidth]{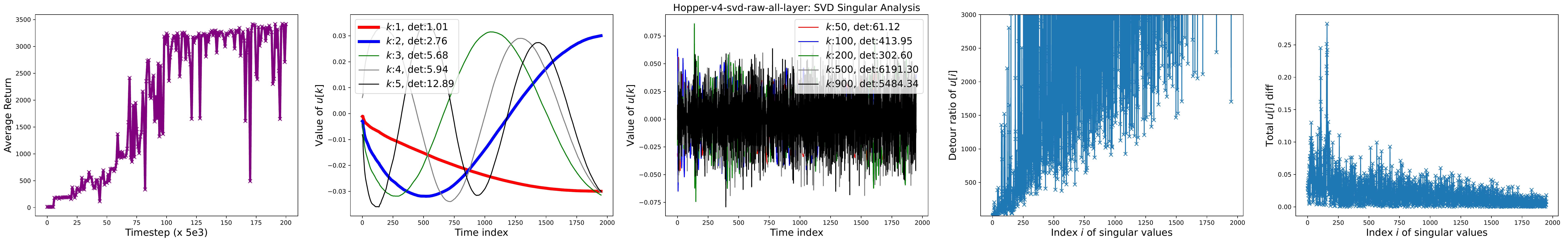}
}
\vspace{-0.2cm}
\subfigure[Walker2d-v4]{
\includegraphics[width=1.0\textwidth]{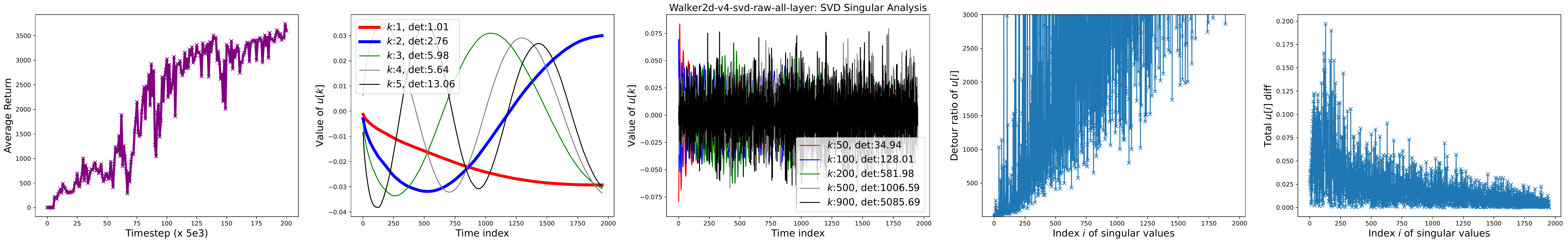}
}
\vspace{-0.2cm}
\subfigure[Ant-v4]{
\includegraphics[width=1.0\textwidth]{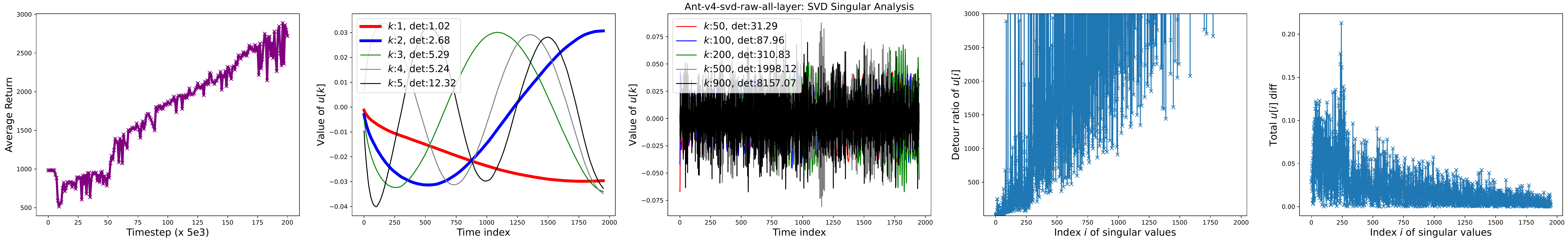}
}
\end{center}
\vspace{-0.3cm}
\caption{Full results of empirical investigation on SVD left unitary matrix of policy learning path in MuJoCo environments.
}
\label{figure:full_svd_u_mujoco}
\end{figure*}

\begin{figure*}
\begin{center}
\vspace{-0.3cm}
\subfigure[cartpole-swingup]{
\includegraphics[width=1.0\textwidth]{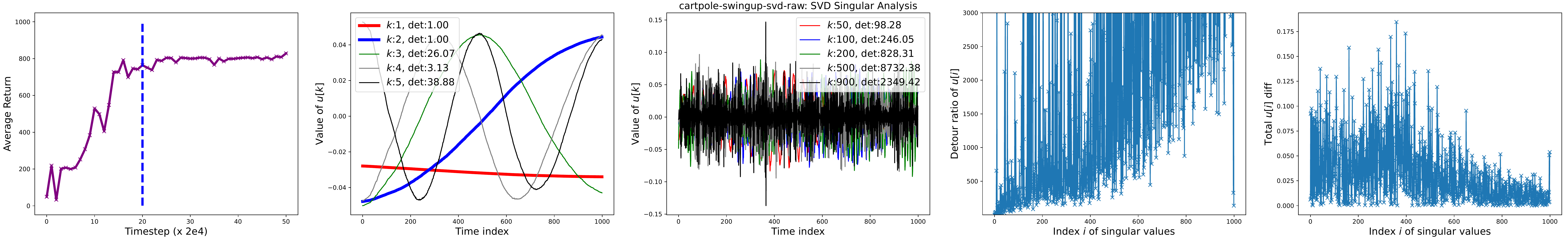}
}
\vspace{-0.2cm}
\subfigure[walker-walk]{
\includegraphics[width=1.0\textwidth]{figs/rad_all_layer_svd_singular_analysis_raw_walker-walk_20221224.pdf}
}
\end{center}
\vspace{-0.3cm}
\caption{Full results of empirical investigation on SVD left unitary matrix of policy learning path in DMC environments.
}
\label{figure:full_svd_u_dmc}
\end{figure*}

\clearpage
\subsection{Empirical Investigation on Temporal SVD Reconstruction of DRL Policies}
\label{appsec:additional_results_svd_recon}

\begin{table}[h]
  \caption{Statistics for Temporal SVD reconstruction of RAD policies in \textbf{cartpole-wingup (the first period)} with varying number of major dimensions kept. 
  }
  \label{table:svd_recon_stats_cw_p0}
\begin{center}
\begin{small}
\begin{sc}
  \scalebox{0.9}{
  \begin{tabular}{c|c|c|c|c}
    \toprule 
    No. of major dim. & AVG($\{\Delta_{\text{R}}(\theta_i)\}_i$) & AVG($\{|\Delta_{\text{R}}(\theta_i)|\}_i$) & max($\{\Delta_{\text{R}}(\theta_i)\}_i$) & min($\{\Delta_{\text{R}}(\theta_i)\}_i$) \\
    \midrule
    1 & 24.24 $\pm$ 65.10 & 43.83 $\pm$ 53.90 & 313.15 & -85.48 \\
    \midrule
    2 & 26.68 $\pm$ 61.70 & 37.45 $\pm$ 55.83 & 278.79 & -71.03 \\
    \midrule
    4 & 14.47 $\pm$ 50.29 & 34.78 $\pm$ 39.10 & 168.46 & -106.50 \\
    \midrule
    8 & 13.86 $\pm$ 45.97 & 32.91 $\pm$ 34.96 & 157.25 & -94.24 \\
    \midrule
    16 & 15.06 $\pm$ 38.30 & 27.69 $\pm$ 30.45 & 125.26 & -52.95 \\
    \midrule
    32 & 8.98 $\pm$ 41.06 & 29.47 $\pm$ 29.97 & 129.39 & -72.51 \\
    \midrule
    64 & 2.15 $\pm$ 41.69 & 28.19 $\pm$ 30.78 & 125.08 & -132.06 \\
    \midrule
    128 & -2.04 $\pm$ 45.76 & 27.95 $\pm$ 36.30 & 120.36 & -202.92 \\   
    \bottomrule
  \end{tabular}
}
\end{sc}
\end{small}
\end{center}
\vskip -0.1in
\end{table}

\begin{table}[h]
  \caption{Statistics for Temporal SVD reconstruction of RAD policies in \textbf{cartpole-wingup (the second period)} with varying number of major dimensions kept. 
  }
  \label{table:svd_recon_stats_cw_p1}
\begin{center}
\begin{small}
\begin{sc}
  \scalebox{0.9}{
  \begin{tabular}{c|c|c|c|c}
    \toprule 
    No. of major dim. & AVG($\{\Delta_{\text{R}}(\theta_i)\}_i$) & AVG($\{|\Delta_{\text{R}}(\theta_i)|\}_i$) & max($\{\Delta_{\text{R}}(\theta_i)\}_i$) & min($\{\Delta_{\text{R}}(\theta_i)\}_i$) \\
    \midrule
    1 & 18.10 $\pm$ 41.86 & 21.01 $\pm$ 40.47 & 265.96 & -12.61 \\
    \midrule
    2 & 8.37 $\pm$ 24.39 & 9.70 $\pm$ 23.89 & 167.57 & -13.24 \\
    \midrule
    4 & 4.71 $\pm$ 21.28 & 10.07 $\pm$ 19.33 & 129.82 & -54.63 \\
    \midrule
    8 & 5.03 $\pm$ 24.63 & 8.28 $\pm$ 23.74 & 164.32 & -53.45 \\
    \midrule
    16 & 4.42 $\pm$ 18.25 & 6.96 $\pm$ 17.44 & 123.94 & -27.75 \\
    \midrule
    32 & 5.48 $\pm$ 35.52 & 10.54 $\pm$ 34.36 & 238.01 & -77.67 \\
    \midrule
    64 & 3.41 $\pm$ 18.15 & 7.35 $\pm$ 16.94 & 116.21 & -32.73 \\
    \midrule
    128 & 3.02 $\pm$ 11.10 & 5.51 $\pm$ 10.09 & 65.10 & -31.63 \\
    \bottomrule
  \end{tabular}
}
\end{sc}
\end{small}
\end{center}
\vskip -0.1in
\end{table}

\begin{table}[h]
  \caption{Statistics for Temporal SVD reconstruction of RAD policies in \textbf{finger-spin (the first period)} with varying number of major dimensions kept. 
  }
  \label{table:svd_recon_stats_fs_p0}
\begin{center}
\begin{small}
\begin{sc}
  \scalebox{0.9}{
  \begin{tabular}{c|c|c|c|c}
    \toprule 
    No. of major dim. & AVG($\{\Delta_{\text{R}}(\theta_i)\}_i$) & AVG($\{|\Delta_{\text{R}}(\theta_i)|\}_i$) & max($\{\Delta_{\text{R}}(\theta_i)\}_i$) & min($\{\Delta_{\text{R}}(\theta_i)\}_i$) \\
    \midrule
    1 & -61.70 $\pm$ 94.46 & 62.70 $\pm$ 93.80 & 10.00 & -393.67 \\
    \midrule
    2 & 2.28 $\pm$ 13.87 & 7.48 $\pm$ 11.91 & 70.00 & -15.67 \\
    \midrule
    4 & 0.77 $\pm$ 18.79 & 8.96 $\pm$ 16.54 & 51.33 & -106.33 \\
    \midrule
    8 & 0.09 $\pm$ 12.15 & 6.49 $\pm$ 10.27 & 28.33 & -62.67 \\
    \midrule
    16 & -5.06 $\pm$ 25.28 & 9.07 $\pm$ 24.13 & 24.33 & -158.00 \\
    \midrule
    32 & -0.83 $\pm$ 7.45 & 5.77 $\pm$ 4.80 & 20.00 & -20.67 \\
    \midrule
    64 & -0.24 $\pm$ 9.50 & 5.15 $\pm$ 7.99 & 18.67 & -52.00 \\
    \midrule
    128 & 1.55 $\pm$ 10.02 & 5.91 $\pm$ 8.24 & 52.33 & -22.33 \\
    \bottomrule
  \end{tabular}
}
\end{sc}
\end{small}
\end{center}
\vskip -0.1in
\end{table}

\begin{table}
  \caption{Statistics for Temporal SVD reconstruction of RAD policies in \textbf{finger-spin (the second period)} with varying number of major dimensions kept. 
  }
  \label{table:svd_recon_stats_fs_p1}
\begin{center}
\begin{small}
\begin{sc}
  \scalebox{0.9}{
  \begin{tabular}{c|c|c|c|c}
    \toprule 
    No. of major dim. & AVG($\{\Delta_{\text{R}}(\theta_i)\}_i$) & AVG($\{|\Delta_{\text{R}}(\theta_i)|\}_i$) & max($\{\Delta_{\text{R}}(\theta_i)\}_i$) & min($\{\Delta_{\text{R}}(\theta_i)\}_i$) \\
    \midrule
    1 & -10.68 $\pm$ 26.53 & 19.89 $\pm$ 20.54 & 45.33 & -90.00 \\
    \midrule
    2 & -5.94 $\pm$ 30.71 & 11.67 $\pm$ 29.02 & 26.67 & -202.00 \\
    \midrule
    4 & -0.95 $\pm$ 10.37 & 6.05 $\pm$ 8.48 & 23.00 & -51.00 \\
    \midrule
    8 & 0.03 $\pm$ 10.39 & 6.04 $\pm$ 8.46 & 26.00 & -54.33 \\
    \midrule
    16 & -0.81 $\pm$ 10.20 & 5.81 $\pm$ 8.42 & 27.33 & -47.00 \\
    \midrule
    32 & -0.47 $\pm$ 10.63 & 5.99 $\pm$ 8.80 & 28.33 & -53.67 \\
    \midrule
    64 & -4.90 $\pm$ 29.45 & 9.85 $\pm$ 28.19 & 24.67 & -197.00 \\
    \midrule
    128 & -16.57 $\pm$ 69.42 & 22.17 $\pm$ 67.84 & 23.00 & -407.67 \\
    \bottomrule
  \end{tabular}
}
\end{sc}
\end{small}
\end{center}
\vskip -0.1in
\end{table}

\begin{table}[t]
  \caption{Statistics for Temporal SVD reconstruction of RAD policies in \textbf{walker-walk (the first period)} with varying number of major dimensions kept. 
  }
  \label{table:svd_recon_stats_ww_p0}
\begin{center}
\begin{small}
\begin{sc}
  \scalebox{0.9}{
  \begin{tabular}{c|c|c|c|c}
    \toprule 
    No. of major dim. & AVG($\{\Delta_{\text{R}}(\theta_i)\}_i$) & AVG($\{|\Delta_{\text{R}}(\theta_i)|\}_i$) & max($\{\Delta_{\text{R}}(\theta_i)\}_i$) & min($\{\Delta_{\text{R}}(\theta_i)\}_i$) \\
    \midrule
    1 & 7.20 $\pm$ 116.51 & 84.63 $\pm$ 80.40 & 285.06 & -298.17 \\
    \midrule
    2 & 61.12 $\pm$ 82.60 & 72.36 $\pm$ 72.96 & 347.76 & -58.92 \\
    \midrule
    4 & 41.79 $\pm$ 86.74 & 67.88 $\pm$ 68.29 & 283.30 & -211.50 \\
    \midrule
    8 & 32.80 $\pm$ 77.18 & 55.46 $\pm$ 62.91 & 261.98 & -144.36 \\
    \midrule
    16 & 25.53 $\pm$ 82.42 & 62.40 $\pm$ 59.59 & 233.16 & -177.25 \\
    \midrule
    32 & 24.22 $\pm$ 86.47 & 61.09 $\pm$ 65.81 & 255.32 & -262.19 \\
    \midrule
    64 & 20.31 $\pm$ 77.88 & 51.96 $\pm$ 61.47 & 298.37 & -225.65 \\
    \midrule
    128 & 12.57 $\pm$ 74.66 & 56.62 $\pm$ 50.27 & 240.83 & -156.34 \\
    \bottomrule
  \end{tabular}
}
\end{sc}
\end{small}
\end{center}
\vskip -0.1in
\end{table}

\begin{table}[t]
  \caption{Statistics for Temporal SVD reconstruction of RAD policies in \textbf{walker-walk (the second period)} with varying number of major dimensions kept. 
  }
  \label{table:svd_recon_stats_ww_p1}
\begin{center}
\begin{small}
\begin{sc}
  \scalebox{0.9}{
  \begin{tabular}{c|c|c|c|c}
    \toprule 
    No. of major dim. & AVG($\{\Delta_{\text{R}}(\theta_i)\}_i$) & AVG($\{|\Delta_{\text{R}}(\theta_i)|\}_i$) & max($\{\Delta_{\text{R}}(\theta_i)\}_i$) & min($\{\Delta_{\text{R}}(\theta_i)\}_i$) \\
    \midrule
    1 & 15.03 $\pm$ 52.61 & 38.35 $\pm$ 39.02 & 155.56 & -149.09 \\
    \midrule
    2 & 21.23 $\pm$ 37.68 & 30.37 $\pm$ 30.80 & 145.45 & -37.95 \\
    \midrule
    4 & 21.81 $\pm$ 34.98 & 28.14 $\pm$ 30.12 & 113.82 & -25.32 \\
    \midrule
    8 & 14.12 $\pm$ 44.51 & 31.12 $\pm$ 34.81 & 134.84 & -163.47 \\
    \midrule
    16 & 6.16 $\pm$ 46.68 & 32.26 $\pm$ 34.30 & 100.79 & -169.74 \\
    \midrule
    32 & 1.55 $\pm$ 62.75 & 35.26 $\pm$ 51.93 & 101.41 & -266.15 \\
    \midrule
    64 & 11.91 $\pm$ 38.39 & 28.99 $\pm$ 27.85 & 125.69 & -75.02 \\
    \midrule
    128 & 14.11 $\pm$ 39.72 & 28.39 $\pm$ 31.15 & 124.47 & -116.62 \\
    \bottomrule
  \end{tabular}
}
\end{sc}
\end{small}
\end{center}
\vskip -0.1in
\end{table}

\end{document}